\pdfoutput=1
\documentclass{article}
\usepackage{spconf,amsmath,graphicx}
\usepackage{float}
\usepackage[dvipsnames]{xcolor}
\usepackage{soul}
\usepackage{tabularx}
\usepackage{multirow}
\usepackage{placeins}
\usepackage{pgfplots}
\pgfplotsset{compat=newest}
\usepackage{subcaption}
\usepackage{stfloats}
\usepackage{hyperref}
\usepackage{todonotes}

\title{Smart Speech Segmentation using acousto-linguistic features\\ with look-ahead}

\name{
\begin{tabular}{@{}c@{}}
Piyush Behre\qquad 
Naveen Parihar\qquad 
Sharman Tan\qquad 
Amy Shah\qquad
Eva Sharma\\
Geoffrey Liu\qquad
Shuangyu Chang\qquad 
Hosam Khalil\qquad
Chris Basoglu\qquad
Sayan Pathak\qquad
\end{tabular}
}
\address{Microsoft Corporation}

\begin{document}

\maketitle
\begin{abstract}

Segmentation for continuous Automatic Speech Recognition (ASR) has traditionally used silence timeouts or voice activity detectors (VADs), which are both limited to acoustic features. This segmentation is often overly aggressive, given that people naturally pause to think as they speak. Consequently, segmentation happens mid-sentence, hindering both punctuation and downstream tasks like machine translation for which high-quality segmentation is critical. Model-based segmentation methods that leverage acoustic features are powerful, but without an understanding of the language itself, these approaches are limited. We present a hybrid approach that leverages both acoustic and language information to improve segmentation. Furthermore, we show that including one word as a look-ahead boosts segmentation quality. On average, our models improve segmentation-F$_{0.5}$ score by 9.8\% over baseline. We show that this approach works for multiple languages. For the downstream task of machine translation, it improves the translation BLEU score by an average of 1.05 points.

\end{abstract}
\begin{keywords}
Speech recognition, audio segmentation, decoder segmentation, continuous recognition
\end{keywords}
\section{Introduction}
\label{sec:intro}

As Automatic Speech Recognition (ASR) quality has improved, the focus has gradually shifted from short utterance scenarios such as Voice Search and Voice Assistants to long utterance scenarios such as Voice Typing and Meeting Transcription. In the short utterance scenarios, speech end-pointing is important for user perceived latency and user experience. Voice Search and Voice Assistants are scenarios where the primary goal is task completion and elements of written form language such as punctuation are not as critical. The output of ASR is rarely revisited after task completion. For long-form scenarios, the primary goal is to generate highly readable well formatted transcription. Voice Typing aims to replace typing with keyboard for important tasks such as typing e-mails or documents, which are more ``permanent" than search queries. Punctuation and capitalization become as important as recognition errors.

Recent research has demonstrated that ASR models suffer from several problems in the context of long-form utterances, such as lack of generalization from short to long utterances \cite{narayanan2019recognizing} and high WER and deletion errors \cite{chiu2021rnn,lu2021input,wang2022vadoi}.
The common practice in the context of long-form ASR is to segment the input stream.
Segmentation quality is critical for optimal WER and punctuation, which is in turn critical for readability \cite{shugrina2010formatting}.
Furthermore, segmentation directly impacts downstream tasks such as machine translation.
Prior works have demonstrated that improvements in segmentation and punctuation lead to significant BLEU gains in machine translation \cite{cho2017nmt,behre2022streaming}.

Conventionally, simple silence timeouts or voice activity detectors (VADs) have been used to determine segment boundaries \cite{ramirez2007voice,yoshimura2020end}.
Over the years, researchers have taken more complex and model-based approaches to predicting end-of-segment boundaries \cite{shriberg2000prosody,ali2018innovative,hou2020segment}.
However, a clear drawback of VAD and many such model-based approaches is that they leverage only acoustic information, foregoing potential gains from incorporating semantic information from text \cite{li2021long}.
Many works have addressed this issue in end-of-query prediction, combining the prediction task with ASR via end-to-end models \cite{maas2018combining, chang2019joint, li2020towards, hwang2020end, lu2022endpoint}.
Similarly, \cite{huang2022e2e} leveraged both acoustic and textual features via an end-to-end segmenter for long-form ASR.

Our main contributions are as follows:
\begin{itemize}
    \item We demonstrate that linguistic features improve decoder segmentation decisions
    \item We use look-ahead to further improve segmentation decisions by leveraging more surrounding context
    \item We extend our approach to other languages and establish BLEU score gains on the downstream task of machine translation
\end{itemize}

\section{Methods}
\label{sec:methods}

\subsection{Models}
We describe three end-pointing techniques, each progressively improving upon the previous. A key contribution of this paper is introducing an RNN-LM in the segmentation decision-making process, which becomes even more powerful when using a look-ahead. Once segments are produced, they continue through a punctuation stage, where a transformer-based punctuation model punctuates each segment. This punctuation model is fixed for all following setups.

\begin{figure*}[t]
\centering
\includegraphics[width=\linewidth]{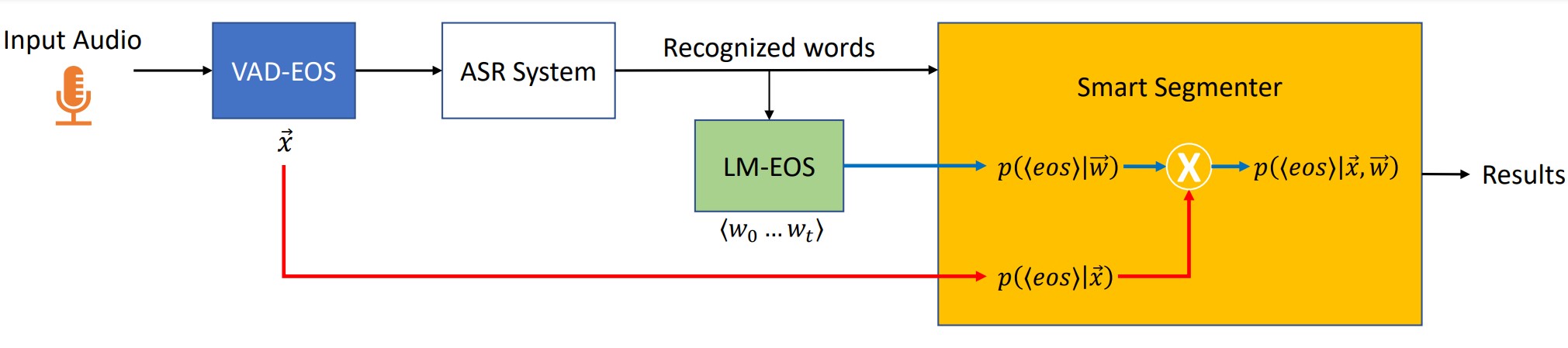}
\caption{Flow chart illustrating hybrid segmentation setup incorporating decisions from VAD-EOS and LM-EOS models. \textbf{\textit{x}} represents LFB-80 features, while \textbf{\textit{w}} represents word embeddings. }
\label{fig:flowchart}
\end{figure*}

\subsubsection{Acoustic/prosodic-only signals (v1)}
In this baseline system, the segmentation decisions are based on a pre-defined threshold for silence. Typically, the default threshold used in such systems is 500ms. This threshold may vary by locale, given that speech rate as well as the frequency and duration of pauses may vary from language to language. This threshold may also vary by scenario. For instance, people tend to speak faster in conversations compared to dictation, so the optimal silence-based timeout threshold may be higher for dictation compared to conversational scenarios like meeting transcription. In addition to silence threshold, the system uses VAD models, which produces better speech end-pointing compared to a simple silence-based timeout approach \cite{ramirez2007voice,yoshimura2020end}.

\subsubsection{Acousto-linguistic signals (v2)}
In natural speech, people often pause disproportionately. Thus, an aggressive v1 setup would result in overly aggressive end-pointing. In the v2 setup, we introduce a language model to offer a second opinion based on linguistic features. We call this an LM-EOS (Language Model – End of Segment predictor) model, as shown in Fig \ref{fig:flowchart}. Since the v2 setup incorporates both acoustic and linguistic features in decision-making, it avoids obvious error cases from v1.

\subsubsection{Acousto-linguistic signals with look-ahead (v3)}
In v2, LM-EOS only has access to left context when predicting end-of-segment boundaries. As prior work has established, this setup is severely limiting for punctuation tasks, where the right context is important for optimal punctuation quality \cite{behre2022streaming}. Therefore, in the v3 setup we incorporate the right context in LM-EOS predictions.

\subsection{Model Training}
Our VAD follows prior works which have extensively covered VAD implementation details \cite{ramirez2007voice}. Here, we describe LM-EOS training in detail. First, let us establish the goal for these models.

\subsubsection{LM-EOS model with no look-ahead}
This model used in v2 is trained to predict whether the input sequence is a valid end of a sentence or not, only looking at the past. As illustrated in the examples below, only looking at the left context to predict can be quite limiting. To train this model, we use the Open Web Text corpus \cite{Gokaslan2019OpenWeb}, and splitting the data into rows with one sentence per row. Each sentence must end in a period or a question mark. We discard any sentences containing punctuation other than periods, commas, and question marks. We then normalize the rows into the spoken form using a WFST (Weighted Finite State Transducers)-based text normalization tool. The LM-EOS model should predict end-of-segment ($\langle eos \rangle$) for every one of the rows, as each row is a sentence. To balance this set of sentences with countercases, we take each sentence and delete the last word. For each of these modified sequences, the model will be trained to predict non-EOS. Examples of the resulting training sequences are illustrated in \ref{tab:v2_training}.

\begin{table}[!hb]
  \centering
  \begin{tabular}{|l|l|l|}
    \hline
    \textbf{Id} & \textbf{Input} & \textbf{Output}\\
    \hline
A1 & how is the weather in seattle & O O O O O eos\\
A2 & how is the weather in & O O O O O\\
B1 & i’m new in town & O O O eos\\
B2 & i’m new in & O O O\\
C1 & wake me up at noon tomorrow & O O O O O eos\\
C2 & wake me up at noon & O O O O O\\
    \hline
  \end{tabular}
  \caption{V2 training data sample rows}
  \label{tab:v2_training}
\end{table}

\begin{table}[!ht]
  \centering
  \begin{tabular}{|l|l|l|}
    \hline
    \textbf{Id} & \textbf{Input} & \textbf{Output}\\
    \hline
A3 & how is the weather in seattle i’m & O O O O O eos O\\
B3 & i’m new in town wake & O O O eos O\\
C3 & wake me up at noon how & O O O O O eos O\\
    \hline
  \end{tabular}
  \caption{V2 training data sample rows}
  \label{tab:v3_training}
\end{table}

We include examples C1 and C2 to highlight an important shortcoming in our v2 training data preparation. The v2 model would not predict $\langle eos \rangle$ for example C2, even though C2 is a perfectly valid sentence. However, we observe that with enough training data, the model still learns that C2 is a valid sentence on its own.

\subsubsection{LM-EOS model with 1-word look-ahead}
The previous training setup can be extended to predict output tags with a one-word delay. This ensures that the model takes the next word into account when making its segmentation decision. To generate the training data for this model, in addition to the data in Table \ref{tab:v2_training}, we add corresponding variants with one word picked from the beginning of the next sentence, as shown in Table \ref{tab:v3_training}.

This model can be further extended to explore 2-word, 3-word, 4-word look-ahead. As we incorporate more right context, the LM-EOS accuracy increases, but so does the decoder latency. Because latency is critical for real-time ASR scenarios, we focus on the 1-word look-ahead setup in our experiments.

\subsection{Experiments}
Our VAD-EOS model is a 3-layer LSTM model with 64 hidden units. All the LM-EOS models are one-layer LSTMs with 1024 hidden units, an embedding size of 256, and a vocabulary size of 250k. These models are trained per locale, with and without the one-word look-ahead. They are trained until convergence.

\newcolumntype{Y}{>{\centering\arraybackslash}X}
\begin{table*}[!hb]
  \centering
  \begin{tabularx}{0.7\textwidth}{|ccYYYY|} 
    \hline
    \multirow{2}{*}{\textbf{Test Set}} &
    \multirow{2}{*}{\centering \textbf{Model}} &
    \multicolumn{3}{c}{\textbf{Segmentation}} &\\
    & & P & R & F$_{0.5}$ & F$_{0.5}$-gain\\
    \hline
    \multirow{2}{*}{\textit{Dictation-100}}
    & {v1} & 0.60 & 0.81 & 0.63 &\\
    & {v2} & 0.70 & 0.78 & 0.71 & 12.7\%\\
    & {v3} & 0.71 & 0.81 & 0.73 & 15.9\%\\
    \hline
    \multirow{2}{*}{\textit{NPR-76}}
    & {v1} & 0.76 & 0.81 & 0.77 &\\
    & {v2} & 0.79 & 0.83 & 0.80 & 3.9\%\\
    & {v3} & 0.82 & 0.82 & 0.82 & 6.5\%\\
    \hline
    \multirow{2}{*}{\textit{EP-100}}
    & {v1} & 0.59 & 0.74 & 0.61 &\\
    & {v2} & 0.63 & 0.74 & 0.65 & 6.6\%\\
    & {v3} & 0.64 & 0.77 & 0.66 & 8.2\%\\
    \hline
    \multirow{2}{*}{\textit{Earnings-10}}
    & {v1} & 0.69 & 0.77 & 0.70 &\\
    & {v2} & 0.73 & 0.78 & 0.74 & 5.7\%\\
    & {v3} & 0.75 & 0.79 & 0.76 & 8.5\%\\
    \hline
  \end{tabularx}
  \caption{LM-EOS with one-word look-ahead (v3) shows the highest segmentation-F$_{0.5}$ gain across scenarios.}
  \label{tab:results_segmentation}
\end{table*}

\subsubsection{Test sets}
We evaluate performance across various scenarios using both public and internal test sets.
\newline\newline
\noindent
\textbf{NPR-76} \cite{npr}: 20 hours of test data from 76 transcribed NPR Podcast episodes.

\noindent
\textbf{EP-100} \cite{europar}: This dataset contains 100 English sessions scraped from European Parliament Plenary videos. It also contains human-labeled translations into German.

\noindent
\textbf{EP-locale} \cite{europar}: This collection of datasets also contains sessions scraped from European Parliament Plenary videos, but across different locales. The collection has 5 locale-specific sets and their corresponding human-labeled translation into English.

\noindent
\textbf{Earnings-10}: 10 hours of earnings call transcription data from the MAEC corpus \cite{li2020maec}

\noindent
\textbf{Dictation-100}: This internal set consists of 100 utterances with human transcriptions.

\subsubsection{Metrics}
We compute segmentation-F$_{0.5}$ as the primary metric, with a higher weight given to precision than recall. This is in line with our users preferring under-segmentation over over-segmentation.
\noindent
For the European Parliament sets, we also compute source-BLEU as well as translation-BLEU scores against the corresponding human-transcribed ground truth sets.

\section{Results}
\label{sec:results}

Our baseline system suffers from over-segmentation problem as indicated by recall being much higher than precision.
Our user studies indicate that precision is much more important than recall; users prefer the system only punctuating when it is confident. Thus, we focus on F$_{0.5}$, weighing precision higher than recall.

Table \ref{tab:results_segmentation} presents the segmentation-F$_{0.5}$ scores across the four test sets. Firstly, the v2 model (LM-EOS model with no look ahead) improves over the baseline across all the datasets. This supports our hypothesis that introducing linguistic features can help segmentation decisions. Furthermore, v3 (LM-EOS model with one-word look ahead) brings additional gains over v2, indicating that using look-ahead to leverage right context is important for LM-EOS to perform well.

It is worth noting that the impact of linguistic features varies across the datasets. The problem of over-segmentation is especially apparent in the dictation task compared to the others. This is as expected, because dictation or voice typing generally involves much more pausing and thinking compared to more natural or prepared speech as in NPR podcasts or earnings calls. For the NPR-76 set, v3 achieves a complete balance between precision and recall.

Table \ref{tab:results_translation} presents results from the translation task for 6 locales. Each locale indicates the language of the original source audio file. The non-English audio files are transcribed and then translated into English. English source files are translated into German. Finally, BLEU scores are computed, matching the machine translation outputs against corresponding human-translated reference texts. The results demonstrate that not only is our hybrid approach effective across locales, but it also significantly boosts performance on the downstream task of machine translation.

\begin{table}[!t]
  \centering
  \begin{tabularx}{0.4\textwidth}{|ccYY|}
    \hline
    {\textbf{Locale}} & {\textbf{Model}} &
    {\centering \textbf{BLEU}} & {\textbf{Gain}}\\
    \hline
    \multirow{3}{*}{en-GB}
    & v1 & 35.5 & \\
    & v2 & 36 & +0.5\\
    & v3 & 36.9 & +1.4\\
    \hline
    \multirow{3}{*}{es-ES}
    & v1 & 41.4 & \\
    & v2 & 41.6 & +0.2\\
    & v3 & 41.8 & +0.4\\
    \hline
    \multirow{3}{*}{el-GR}
    & v1 & 34.5 & \\
    & v2 & 35.1 & +0.7\\
    & v3 & 35.3 & +0.8\\
    \hline
    \multirow{3}{*}{fr-FR}
    & v1 & 38.4 & \\
    & v2 & 39.1 & +0.7\\
    & v3 & 39.6 & +1.2\\
    \hline
    \multirow{3}{*}{it-IT}
    & v1 & 44.7 & \\
    & v2 & 45.7 & +1.0\\
    & v3 & 46 & +1.3\\
    \hline
    \multirow{3}{*}{ro-RO}
    & v1 & 40.3 & \\
    & v2 & 40.9 & +0.6\\
    & v3 & 41.5 & +1.2\\
    \hline
  \end{tabularx}
  \caption{On EP-locale sets, v3 improves BLEU by 0.4-1.4 pt. \\en-GB translated to de-DE, other locales translated to en-GB.}
  \label{tab:results_translation}
\end{table}

\section{Discussion}
\label{sec:discussion}

We establish that introducing the linguistic features significantly improves decoder segmentation decisions. Our preliminary work on LM-EOS models with look-ahead demonstrates that leveraging limited right context is a powerful way to maximize the gains from incorporating linguistic features. We focused on LSTMs due to latency constraints of the system. In the future, we plan to explore transformer-based LM-EOS with much larger look-ahead values to capture even more right context. While this has latency implications, it is possible that with a more powerful LM-EOS, we may be able to rely a lot more on this model even in the absence of VAD signals to endpoint sooner. At the very least, it could benefit near-real-time or non-real-time use cases like Voicemail Transcriptions. 

Our current approach combines the decisions from two separate models. The VAD-EOS model is language independent, while the LM-EOS model is trained per language. This allows us to train and improve them separately, which is critical considering the sparsity of publicly available audio and corresponding human-punctuated transcriptions.
This allows for easier insertion of custom vocabulary into LM-EOS or early end-pointing based on detection of certain keywords or commands.

\section{Conclusion}
\label{sec:format}

In this paper, we explored techniques that can improve speech segmentation. We demonstrated that linguistic features improve decoder segmentation decisions and look-ahead is a good way to leverage surrounding context in this decision making. We demonstrated its effectiveness on a wide range of scenarios. The method is most effective for \textit{human2machine} scenarios like Dictation where users pause and think a lot. We see a segmentation-F$_{0.5}$ gain of 15.9\% for this. The technique nevertheless proves effective for other \textit{human2human} scenarios as well. We prove that the technique can work well for other languages. We establish the benefit of improved segmentation for the downstream task of machine translation, measured by BLEU score gains of 0.4-1.4 across languages. In the future, we will explore transformer-based LM-EOS with longer look-ahead particularly targeting non-real-time use cases.

\FloatBarrier

\vfill\pagebreak

\bibliographystyle{IEEEbib}
\bibliography{refs}

\end{document}